\documentclass[final]{ifacconf}  
\usepackage{natbib}
\usepackage{graphicx}
\usepackage{array}
\usepackage{multirow}
\usepackage{tabularx}
\usepackage{epstopdf}
\usepackage{graphics} 

\usepackage[]{units}
\usepackage{pgfplots}
\usepackage{grffile}
\pgfplotsset{compat=newest}
\usepgfplotslibrary{patchplots}

\usepackage{paralist}
\usepackage{mathtools}

\usepackage{xcolor}
\usepackage{siunitx}
\graphicspath{ {Pictures/} }
\usepackage{epsfig}
\usepackage{amsmath} 
\usepackage{amssymb}  
\usepackage{mathtools}
\usepackage{bm}
\graphicspath{ {Pictures/} }
\DeclareGraphicsExtensions{.pdf,.eps}

\usepackage[acronym,nomain,nonumberlist]{glossaries}
\newacronym{mav}{MAV}{Micro Aerial Vehicle}
\newacronym{nmpc}{NMPC}{Nonlinear Model Predictive Control}
\newacronym{mpc}{MPC}{Model Predictive Control}
\newacronym{panoc}{PANOC}{Proximal Averaged Newton-type method for Optimal Control}

\begin{document}
\begin{frontmatter}

\title{Collision Avoidance for Multiple MAVs using Fast Centralized NMPC\thanksref{footnoteinfo}} 

\thanks[footnoteinfo]{This work has been partially funded by the European Unions Horizon 2020 Research and Innovation Programme under the Grant Agreement No. 730302 SIMS. Corresponding author's e-mail: bjolin@ltu.se}

\author[First]{Bj\"orn Lindqvist} 
\author[First]{Sina Sharif Mansouri} 
\author[Second]{Pantelis Sopasakis} 
\author[First]{George Nikolakopoulos} 

\address[First]{Robotics Team, Department of Computer, Electrical and Space Engineering, Lule\r{a} University of Technology, Lule\r{a} SE-97187, Sweden.}
\address[Second]{School of Electronics, Electrical Engineering and Computer Science (EEECS), Queen's University Belfast and Centre for Intelligent Autonomous Manufacturing Systems (i-AMS), United Kingdom}

\begin{abstract}
This article proposes a novel control architecture using a centralized nonlinear model predictive control (CNMPC) scheme for controlling multiple micro aerial vehicles (MAVs). The control architecture uses an augmented state system to control multiple agents and performs both obstacle and collision avoidance. The optimization algorithm used is OpEn, based on the proximal averaged Newton type method for optimal control (PANOC) which provides fast convergence for non-convex optimization problems. The objective is to perform position reference tracking for each individual agent, while nonlinear constrains guarantee collision avoidance and smooth control signals. To produce a trajectory that satisfies all constraints a penalty method is applied to the nonlinear constraints. The efficacy of this proposed novel control scheme is successfully demonstrated through simulation results and comparisons, in terms of computation time and constraint violations, while are  provided with respect to the number of agents. 
\end{abstract}

\end{frontmatter}

\glsresetall 

\section{Introduction}

\subsection{Background and Motivation}

The deployment of coordinated MAV flights is gaining increasing attention in different 
applications areas, such as coordinated inspection of 
infrastructure~\citep{mansouri2018cooperative} or coordinated aerial 
acrobatics~\citep{mellinger2011minimum}. Such application scenarios call for 
advanced control algorithms that provide high levels of autonomy with integrated 
collision avoidance, among the agents and obstacles in the environment.


A centralized control scheme allows all agents to account for obstacles 
and changes in their environment, as well as the positions of other agents, without 
agent-to-agent broadcasting. Such a control scheme must provide 
satisfactory trajectories for all agents, while keeping the computation time low as
it is required for the control of MAVs. 

Specifically, for a nonlinear model predictive control (NMPC) 
scheme, the computation time is the greatest bottleneck. But provided that the 
computation time is constrained low for multiple MAVs, the customizability of the
NMPC makes it a great scheme for applying a centralized control. In this article, we present
a nonlinear constrained NMPC, based on the proximal averaged Newton 
method for optimal control (PANOC) \citep{panoc2017,small2019aerial,sathya2018embedded} 
that provides obstacle and collision avoidance 
for multiple agents using a penalty method \citep{Hermans:IFAC:2018}, 
while keeping the computation times 
below a desired bound of $\unit[50]{ms}$ \citep{small2019aerial}. 

The problem of path planning \citep{lavalle2006planning} has been 
widely studied as it a key problem in mobile robotics. In the specific 
case for MAVs it comes with its own specific challenges, such as the 
nonlinearity of the system dynamics and the tight runtime requirements, while several approaches have been proposed in the related literature \citep{goerzen2010survey}. 
Popular path planners include the potential fields 
\citep{droeschel2016multilayered}, graph search methods, such as 
dynamic variations of $A^{*}$ \citep{heng2011autonomous} and model 
predictive control \citep{alexis2011switching}. The overarching goal 
of such path planners is to have a reactive control system that can 
provide collision-free paths in constrained environments, while 
accounting for the nonlinear dynamics of the MAV
at a low computation time.

The problem of cooperative path-planning for multiple MAVs can be 
split into centralized and non-centralized (distributed, 
decentralized). 
The distributed and decentralized schemes have been widely used \citep{montenegro2015review, kumaresan2016decentralized,richards2004decentralized,chao2012uav},
where only specific data are shared among agents. 
Specifically in regards to a distributed or decentralized MPC, it is a common practice to either share the predicted trajectories \citep{richards2004decentralized,chao2012uav}
or internally predict the positions of other 
agents using a linear prediction model, to prevent collisions \citep{kamel2017nonlinear}. 
In such decentralized approaches, individual agents solve
lower dimensional optimization problems, thus mitigating the 
heavy load of solving one large centralized problem.
However, according to \cite{negenborn2010intelligence} centralized
formulations are computationally demanding and outperform their
distributed/decentralized counterparts.

A legitimate concern, in the centralized control architectures, is the robustness of the communication of the vehicles to a central 
computing node. However, with more stable and faster wireless networks, such as 5G and edge computing, a centralized control scheme becomes relevant. In such centralized control schemes, the entire position and orientation information is available to a single computing agent, which can make informed decisions towards a collision-free navigation of all vehicles, while this in turn provides a better closed-loop performance. Centralized schemes are also suitable for small MAVs that do not have adequate computing power. 
Instead, such computation can, for example, be offloaded to an edge computing system and wirelessly transmitted to the agents~\citep{varghese2016challenges}. Centralized NMPC schemes have been considered before, such as \citep{erunsal2019nonlinear}.

All the aforementioned cases of NMPC use some form of SQP 
(sequential quadratic programming) to solve the optimization problem. The main 
drawback of SQP lies in the need to solve a quadratic program at each iteration, 
which requires inner iterations, and the requirement to compute and store Jacobian 
matrices. The proposed method uses PANOC \citep{stella2017simple, sathya2018embedded} as 
the solver for the centralized control scheme. 
PANOC uses the same oracle as the projected gradient method \citep{nesterov2018lectures}, involves only simple algebraic
operations and has a low memory and computational footprint. 
PANOC has been shown to successfully fly a single MAV in 
an obstructed environment \citep{small2019aerial}.
%
\subsection{Contributions}
%
The first contribution of this article is a framework for centralized NMPC (CNMPC) of multiple MAVs. This framework provides collision-free paths, while avoiding obstacles. We consider nonlinear dynamical models for every aerial agent and solve a common optimization problem to decide individual control actions and unlike distributed approaches, we optimize over the trajectories of all agents.

Secondly, we conduct several realistic simulations to evaluate the 
proposed CNMPC approach in demanding environments (tight formations
and multiple obstacles). We analyze the efficacy of CNMPC in terms
of the required execution time and show that this scales gracefully 
with the number of agents. We also evaluate the constraint violations
in closed-loop simulations, which are found to be low.

Lastly, we briefly compare the performance of our approach to that of SQP. As it will be demonstrated, the OpEn-based method successfully leads to collision-free paths in all the cases, while the computation time remains within the tight prescribed requirements of $\unit[50]{ms}$ for up to seven agents.

\section{Methodology} \label{sec:methodology}
%
\subsection{MAVs Kinematics} \label{sec:mavkinematic}
\begin{figure}
\centering
  \includegraphics[width=0.8\linewidth]{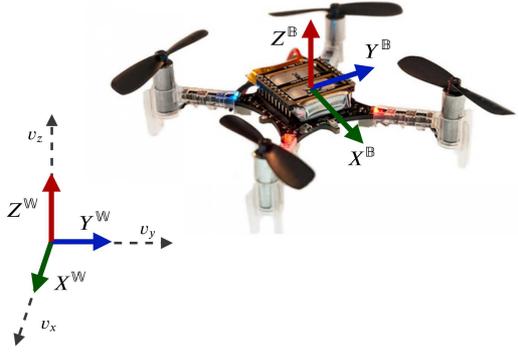}
  \caption{Utilized coordinate frames, where $\mathbb{W}$ and $\mathbb{B}$ denote the world and body coordinate frames respectively.}
  \label{fig:coordinateMAV}
\end{figure}

The MAV coordinate systems are depicted in Figure~\ref{fig:coordinateMAV}, where $(x^\mathbb{B}, y^\mathbb{B}, z^\mathbb{B})$ denote the body-fixed coordinate system, while $(x^\mathbb{W}, y^\mathbb{W}, z^\mathbb{W})$ denote the global coordinate system. In this article the states of all agents are defined in a yaw-compensated global frame of reference.
The six degrees of freedom (DoF) MAV is defined by the set of equations \eqref{eq:mavkinematic}. The full derivation of the adopted model can be found in~(\cite{kamel2017model}). 
\begin{subequations}
\label{eq:mavkinematic}
\begin{align}
        \dot{p}(t) &= v(t) \\ 
        \dot{v}(t) &= R(\phi,\theta) 
        \begin{bmatrix} 0 \\ 0 \\ T \end{bmatrix} + 
        \begin{bmatrix} 0 \\ 0 \\ -g \end{bmatrix} - 
        \begin{bmatrix} A_x & 0 & 0 \\ 0 &  A_y & 0 \\ 0 & 0 & A_z \end{bmatrix} v(t), \\ 
        \dot{\phi}(t) & = \nicefrac{1}{\tau_{\phi}} (K_\phi\phi_{\mathrm{ref}}(t)-\phi(t)), \\ 
        \dot{\theta}(t) & = \nicefrac{1}{\tau_{\theta}} (K_\theta\theta_{\mathrm{ref}}(t)-\theta(t)),
\end{align}
\end{subequations}
where $p=[p_x,p_y,p_z]^\top$ is the position, $v = [v_x,v_y,v_z]^\top$ is the linear velocity in the global frame of reference, and $\phi$ and $\theta \in[-\pi,\pi]$ are the roll and pitch angles along the $x^\mathbb{W}$ and $y^\mathbb{W}$ axes respectively. Moreover, $R(\phi(t),\theta(t)) \in \mathrm{SO}(3)$ is a rotation matrix that describes the attitude in Euler form, with $\phi_{\mathrm{ref}}\in \mathbb{R}$, $\theta_{\mathrm{ref}}\in \mathbb{R}$ and $T\geq 0$ to be the references in roll, pitch and the total thrust generated by the four rotors. The above model assumes that the acceleration depends only on the magnitude and angle of the thrust vector, produced by the motors, as well as the linear damping terms $A_x, A_y, A_z \in \mathbb{R}$ and the gravitational acceleration $g$.

The attitude terms are modeled as a first-order system between the attitude (roll/pitch) and the references $\phi_{\mathrm{ref}}\in \mathbb{R}$, $\theta_{\mathrm{ref}}\in \mathbb{R}$, with gains $K_\phi, K_\theta\in\mathbb{R}$ and time constants $\tau_\phi, \tau_\theta \in \mathbb{R}$. The aforementioned terms model the closed-loop behavior of a low-level controller, which also implies that the MAV is equipped with a lower-level attitude controller that takes thrust, roll and pitch commands and provides motor commands for the MAV such as~\citep{jackson2016rosflight}.   
%

\subsection{Joint Cost Function} \label{sec:obj}
%
Let the state vector of agent \textit{i} be denoted by $x^{(i)} = [p^{(i)}, v^{(i)}, \phi^{(i)}, \theta^{(i)}]^\top$, and the corresponding control action be $u^{(i)}=[T^{(i)},\phi_{\mathrm{ref}}^{(i)},\theta_{\mathrm{ref}}^{(i)}]^\top$. The state variables of all agents can be collected into the vector $x = [x^{(1)}, x^{(2)}, \ldots, x^{(N_a)}]^\top$ where $N_a$ is the number of agents. Likewise, we define the vector of all control actions, $u = [u^{(1)}, u^{(2)}, \ldots, u^{(N_a)}]^\top$.

The system dynamics of all $N_a$ agents are discretized with a sampling time of $T_s=\unit[50]{ms}$ using the forward Euler method to obtain
\begin{equation}
\label{eq:prediction}
    x_{k+1} = f(x_k, u_k).
\end{equation}
This discrete model is used as the \textit{prediction model} of NMPC. This prediction is done with receding horizon e.g., the prediction considers a set number of steps into the future. We denote this as the \textit{prediction horizon}, $N$, of the NMPC. By associating a cost to a configuration of states and inputs at the current time and in the prediction, a nonlinear optimizer is tasked with finding an optimal set of control actions, defined by the cost minimum of this \textit{cost function}. 

Let $x_{k+j{}\mid{}k}^{(i)}$ denote the predicted state of agent $i$ 
at time step $k+j$, produced at the time step $k$. The corresponding
control actions are denoted by $u_{k+j{}\mid{}k}^{(i)}$.
Let us also denote $\bm{x}_{k} = (x_{k+j{}\mid{}k}^{(i)})_{j,i}$ and
$\bm{u}_{k} = (u_{k+j{}\mid{}k}^{(i)})_{j,i}$.
The controller aims to make the state reach the prescribed set points, while delivering smooth control inputs. To that end, we formulate the following cost function:
\begin{multline}
\label{eq:costfunction}
J(\bm{x}_{k}, \bm{u}_{k}; u_{k-1\mid k}) = \sum_{j=0}^{N}\sum_{i=1}^{N_a}   \underbrace{\| x_{\mathrm{ref}}^{(i)}-x_{k+j{}\mid{}k}^{(i)}\|_{Q_x}^2}_\text{State cost} 
\\
+   \underbrace{\| u_{\mathrm{ref}}-u_{k+j{}\mid{}k}^{(i)}\|^2_{Q_u}}_\text{Input cost}
+  \underbrace{\| u_{k+j{}\mid{}k}^{(i)}-u_{k+j-1{}\mid{}k}^{(i)} \|^2 _{Q_{\Delta u}}}_\text{Input smoothness cost},
\end{multline}
where $Q_x\in \mathbb{R}^{8\times8}, Q_u, Q_{\Delta u}\in 
\mathbb{R}^{3\times3}$ are symmetric  positive definite weight matrices for the
states, inputs and input rates respectively. In \eqref{eq:costfunction}, the 
first term denotes the \textit{state cost}, which penalizes deviating from a 
certain state reference $x_{\mathrm{ref}}$. The second term denotes the 
\textit{input cost} that penalizes a deviation from the steady-state input 
$u_{\mathrm{ref}} = [g, 0, 0]$ i.e. the inputs that describe hovering. Finally,
to enforce smooth control actions, a third term is added that penalizes changes
in successive inputs. Note that the first such penalty, $\| 
u_{k{}\mid{}k}^{(i)}-u_{k-1{}\mid{}k}^{(i)} \|^2$, depends on the previous
control action $u_{k-1{}\mid{}k}^{(i)} = u_{k-1}^{(i)}$.

All agents are given the same weight matrices $Q_x, Q_u, Q_{\Delta u}$ and input reference $u_{\mathrm{ref}}$. This approach considers a separate state reference $x_{\mathrm{ref}}$ for each agent, which allows for a path-planning approach where each agent can track an individual set-point reference. The $\Delta u$-terms $u_{k+j|k}-u_{k+j-1|k}$ are also considered for each agent as the computed control inputs will be different for every agent.   
%
\subsection{Input constraints} \label{sec:input_constraints}
With the goal of being applicable to a real MAV, hard bounds on reference angles $\phi_{\mathrm{ref}}, \theta_{\mathrm{ref}}$ must be considered as a low-level controller will only be able to stabilize the attitude within a certain range. Since the thrust of a MAV is limited, such hard bounds must also be applied to the thrust input, \textit{T}. Thus we define bounds on inputs as:
\begin{equation}
u_{\min} \leq u_{k+j\mid k}^{(i)} \leq u_{\max},
\end{equation}
that are considered for all agents in the system, for every step in the prediction. 
\subsection{Obstacle and collision avoidance} \label{sec:const}
%
\subsubsection{Cylindrical Obstacle} 
Cylinders, as well as balls and rectangles, are convenient geometrical
shapes for enveloping arbitrary obstacles in the environment of the 
aerial vehicles. A cylindrical obstacle can be identified by its position, height and radius, that is, the triplet: 
\(
    \xi^{\mathrm{obs}} 
    \coloneqq 
    (
        p^{\mathrm{obs}}, 
        r^{\mathrm{obs}}, 
        l^{\mathrm{obs}})
\).
Following \cite{sathya2018embedded}, we define the 
obstacle avoidance constraint to be:
\begin{multline}\label{eq:cylinderconstraint} 
    h_{\mathrm{cyl}}(p; \xi^{\mathrm{obs}}) {}\coloneqq{} 
    [p_z-p_z^{\mathrm{obs}}+\tfrac{1}{2}l_{\mathrm{obs}}]_+
    \\
    [p_z+p_z^{\mathrm{obs}}+\tfrac{1}{2}l_{\mathrm{obs}}]_+[r_{\mathrm{obs}}^2 - (p_x{-} p_x^{\mathrm{obs}})^2
    \\
    - (p_y{-}p_y^{\mathrm{obs}})^2]_+ = 0,
\end{multline}
where for all $x\in\mathbb{R}$ we define $[x]_+=\max\{0, x\}$.
In other words, the point $p=(p_x,p_y,p_z)$ lies outside the cylinder at hand
so long as \eqref{eq:cylinderconstraint} holds. 
In the NMPC we shall require that the above constraint is satisfied for all 
positions of all agents along the prediction horizon.
Obstacles of more general shapes can be considered in a similar
fashion following \citep{small2019aerial,sathya2018embedded}.
\subsubsection{Collision Avoidance}
In a similar fashion, in order to prevent collisions among agents 
we require that:
\begin{multline}\label{eq:distance} 
    h_{l,i}(\bm{x}_{k}) 
    {}\coloneqq{} 
    \big[p_{z, k+j\mid k}^{i}-p_{z,k+j\mid k}^{l}+L\big]_+
    \big[p_{z, k+j\mid k}^{i}
    \\+p_{z, k+j\mid k}^{l}+L\big]_+
   \big[r_{\mathrm{safety}}^2 - (p_{x,k+j|k}^{(i)}-p_{x,k+j|k}^{(l)})^2 
    \\ - (p_{y,k+j|k}^{(i)}-p_{y,k+j|k}^{(l)})^2\big]_+ 
    {}={} 0,
\end{multline}
for all pairs $(i,l)$ with $i,l=1,\ldots,N_a$ and $i < l$,
where $r_{\mathrm{safety}}$ is the minimum distance that agents $l$
and $i$ should be away from one another. 
In our NMPC formulation, we shall impose the above constraint between all 
agents $i$ and $l$, $i\neq l$ and along the prediction horizon.
Note that it is not recommended for the vehicles to fly one above the 
other, as the top vehicle's wake can have a destabilizing effect on the 
lower vehicle, thus \eqref{eq:distance} imposes a high safety distance above
each agent.

\subsection{Control Input Rate}
We impose a constraint on the successive differences of control actions
so as to prevent an overly aggressive behavior in the control inputs 
$\phi_{\mathrm{ref}}$ and $\theta_{\mathrm{ref}}$, that is
\begin{subequations}
\begin{align}
    |\phi_{\mathrm{ref}, k+j-1{}\mid{}k}^{(i)} - \phi_{\mathrm{ref},k+j{}\mid{}k}^{(i)}| 
    {}\leq{}
    \Delta \phi_{\max},
    \\
    |\theta_{\mathrm{ref}, k+j-1{}\mid{}k}^{(i)} - \theta_{\mathrm{ref},k+j{}\mid{}k}^{(i)}| 
    {}\leq{}
    \Delta \theta_{\max},
\end{align}
\end{subequations}
for $j = 0, \ldots, N-1$ and $i=1,\ldots, N_a$. The above inequality
constraints can be written as equality constraints as follows: 
\begin{subequations}\label{eq:delta_constraints}
\begin{align}
    [\phi_{\mathrm{ref}, k+j-1{}\mid{}k}^{(i)} 
    - \phi_{\mathrm{ref},k+j{}\mid{}k}^{(i)} 
    -\Delta \phi_{\max}]_+ {}={}& 0,
    \\
    [\phi_{\mathrm{ref},k+j{}\mid{}k}^{(i)} 
    -\phi_{\mathrm{ref}, k+j-1{}\mid{}k}^{(i)} 
    -\Delta \phi_{\max}]_+ {}={}& 0,
\end{align}
\end{subequations}
and similarly for $\theta$.


\subsection{NMPC and Embedded Optimization} \label{sec:solver}
The requirements we outlined above lead to the formulation 
of the following model predictive control problem for $N_a$ 
agents and in presence of multiple cylindrical obstacles, 
identified by $\xi^{\mathrm{obs}}_s$ for $s=1,\ldots, N_o$
\begin{subequations}\label{eq:nmpc}
\begin{align}
    \operatorname*{Minimize}_{
        \bm{u}_k, \bm{x}_k
    } \,
    & J(\bm{x}_{k}, \bm{u}_{k}; u_{k-1\mid k})
    \\
    \text{subject to:}\,&
    x_{k+j+1\mid k} = f(x_{k+j\mid k}, u_{k+j\mid k}),\notag\\
    &\quad j=0,\ldots, N-1,
    \\
    &u_{\min} \leq u_{k+j\mid k}^{(i)} \leq u_{\max}, j=0,\ldots, N,
\label{eq:nmpc:input_constraints}
    \\
    &h_{\mathrm{cyl}}(p_{k+j\mid k}^{(i)}; \xi^{\mathrm{obs}}_s) = 0,
     j=0,\ldots, N,\notag
     \\
     &\quad i=1,\ldots, N_a, s=1,\ldots, N_o,
     \\
     & h_{l,i}(\bm{x}_{k}) = 0,  j=0,\ldots, N,\notag
     \\
     &\quad i, l=1,\ldots, N_a, i {}<{} l,
     \\
     &\text{Constraints \eqref{eq:delta_constraints}},
     j=0,\ldots, N,\notag
     \\
     &\quad i=1,\ldots, N_a,
     \\
     &x_{k{}\mid{}k}^{(i)} = x_k^{(i)}, i=1,\ldots, N_a,
     \\
     &u_{k-1{}\mid{}k}^{(i)} = u_{k-1}^{(i)}, i=1,\ldots, N_a.
\end{align}
\end{subequations}
This problem fits into the framework of the open-source solver 
\texttt{OpEn} \citep{open2019}, which generates embedded-ready 
source code written in Rust; a fast programming language that comes with memory safety guarantees. \texttt{OpEn} solves parametric optimization problems of the general form: 
\begin{subequations}\label{eq:open_problem}
\begin{align}
    \operatorname{Minimize}_{z \in Z}\,& \ell(z)
    \\
    \text{subject to:}\,& F(z) = 0,
\end{align}
\end{subequations}
where $U$ is a set on which one can easily compute projections,
$\ell$ is a Lipschitz-differentiable function and $F$ is a vector-valued mapping so that $\|F(u)\|^2$ is a Lipschitz-differentiable function.

In order to write Problem \eqref{eq:nmpc} in the form of Problem 
\eqref{eq:open_problem}, we define the decision variable 
$z=\bm{u}_k$, choose $Z$ to be the rectangle defined by the 
input constraints \eqref{eq:nmpc:input_constraints}, 
eliminate the system dynamics following the single shooting 
approach of \citep{small2019aerial,sathya2018embedded} and 
define $F$ so as to cast all equality constraints.
The quadratic penalty method formulates gauge problems
(referred to as \textit{inner} problems), which have the 
form: 
\(
    \operatorname{Minimize}_{z \in Z} \ell(z) + c\|F(z)\|^2
\),
where $c$ is a positive penalty parameter. The inner problems are solved using PANOC and the penalty parameter is increased in an \textit{outer} iteration loop until 
$\|F(z)\|_\infty$ drops below a specified \textit{infeasibility 
tolerance}.

\section{Simulation Results} \label{sec:results}
\subsection{Simulation Model Parameters and Costs}
For the presented simulations, the corresponding model parameters can be described as in \eqref{eq:mavkinematic} and are chosen as $\tau_{\phi}, \tau_{\theta} = 0.5$, $K_{\phi}, K_{\theta} = 1$, in order to approximately match the response of a low-level controller acting on a MAV. Additionally, $g$ is set to $\unitfrac[9.82]{m}{s^2}$, the control horizon is set to $N = 30$ which, with a sampling time of $\unit[50]{ms}$, implies a prediction of $\unit[1.5]{s}$. The weights in \eqref{eq:costfunction} are chosen as:
\begin{subequations}
\begin{align}
    Q_x =& \operatorname{diag}(5,5,20,3,3,3,8,8), \\
    Q_u =& \operatorname{diag}(5, 10, 10), \\
    Q_{\Delta u} =& \operatorname{diag}(10, 25, 25).
\end{align}
\end{subequations}
The bounds on control inputs are chosen (in SI units) as: 
\begin{equation}
\label{eq:inputconstraints}
u_{\min} = 
    \begin{bmatrix}
    5 \\
    -0.4 \\
    -0.4
    \end{bmatrix} \\,
    u_{\max} = 
    \begin{bmatrix}
    13.5 \\
    0.4 \\
    0.4
    \end{bmatrix},
\end{equation}
Additionally the constraints on change in the input described in \eqref{eq:delta_constraints} are chosen as $\Delta\phi_{\mathrm{max}} = 0.07$ and $\Delta\theta_{\mathrm{max}} = 0.07$.\\

All simulations use four outer/penalty iterations (except stated otherwise) with initial penalty weight 10 and an update factor of 10 for an exponential increase in costs associated with violating the penalty constraints (obstacle, collision, change in input). In the following simulations we chose obstacles with a large $l_{obs}$ to discourage the MAVs to move above the obstacle. 

The state update model is the same as the non-linear discretized prediction model with an addition of a Gaussian noise parameter. This noise represents a general uncertainty in state data, as well as an uncertainty in how the MAVs behave based on a certain input. Adding noise to the state update forces the optimizer to make realistic micro-adjustments to compensate. The noise parameter is generated with a normal Gaussian distribution with a specified mean, $\mu$, and standard deviation, $\sigma^2$ as $\mathcal{N}(\mu, \sigma^2)$  \citep{peebles2001probability}. 
The noise added to each state are the IID (independent and identically distributed) processes 
$\eta_p \sim \mathcal{N}(0, 0.01)$, 
$\eta_v \sim \mathcal{N}(0, 0.005)$ and 
$\eta_{\theta,\phi} \sim \mathcal{N}(0, 0.001)$, 
where $\eta_p$, $\eta_v$, and $\eta_{\theta,\phi}$ 
are the noise added to the position, velocity, and attitude terms respectively.

\subsection{Simultaneous Collision and Obstacle Avoidance}\label{sec:results1}
The task of the MAVs in the following simulations is to take off to a set point. At $\unit[2.5]{s}$ a new reference set-point is given on the opposite side of a cylindrical obstacle with a radius of $\unit[0.8]{m}$ which forces the MAVs to translate past the obstacle while avoiding collisions to arrive at the new set-point.

\begin{figure}
\centering
  \includegraphics[width=\linewidth]{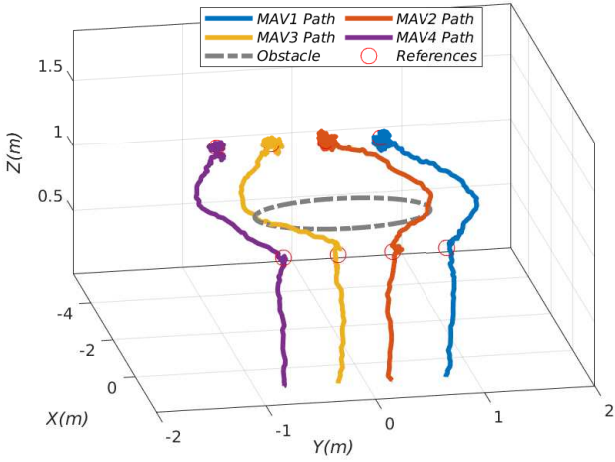}
  \caption{Path of four MAVs avoiding a cylinder with radius 0.8m while keeping a distance of 0.4m.}
  \label{fig:4CylinderPath}
\end{figure}

\begin{figure}
\centering
  \includegraphics[width=\linewidth]{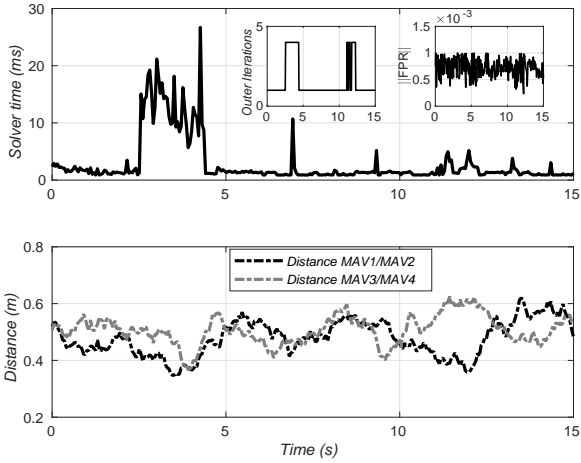}
  \caption{Solver time (top) and distance between agents (bottom) from four-agent simulation. The two insets in the top figure show the number of outer iterations of the penalty algorithm and the norm of the fixed-point residual for the inner solver, which serves as a measure of the quality of the solution.}
  \label{fig:4CylinderData}
\end{figure}

\begin{figure}
\centering
  \includegraphics[width=\linewidth]{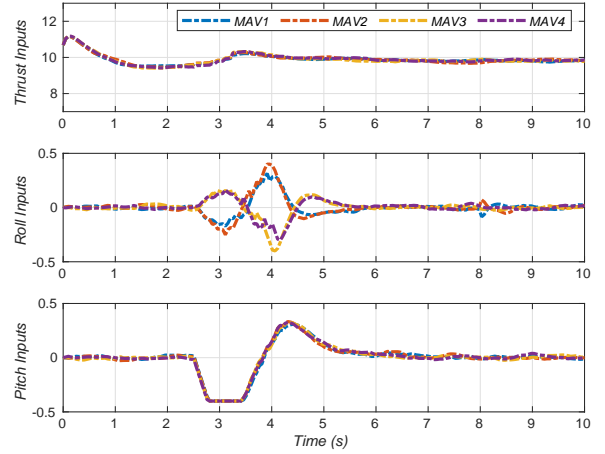}
  \caption{Computed control inputs for four-agent simulation. Thrust (top), Roll Reference (middle), Pitch Reference (top).}
  \label{fig:4CylinderInputs}
\end{figure}

Figure~\ref{fig:4CylinderPath} shows the result by deploying four agents. The application of the collision avoidance constraint is demonstrated by the fact that the trajectory of the outer agents (MAV1 and MAV4) curves out as to satisfy a distance of $\unit[0.4]{m}$ to the inner agents (MAV2 and MAV3). The distance between MAVs can be seen in Figure \ref{fig:4CylinderData}. The safety distance is kept to $\unit[0.4]{m}$ throughout the avoidance phase, except very small violations at $4$ and $\unit[12]{s}$ respectively, with a maximum of $\unit[0.04]{m}$, which is to be expected due to the noisy state update and limited penalty method iterations. The computation time is found in the same figure and spikes up to a maximum of $\unit[27]{ms}$. The insets in the solver time graph shows when the penalty method iterations are added to the computation. Figure \ref{fig:4CylinderInputs} shows the control inputs applied to the state update function. The control inputs display a relatively smooth behavior despite the added noise and momentary constraint violations demonstrating the application of \eqref{eq:delta_constraints}, and is kept within the control input constraints from \eqref{eq:inputconstraints}. 

\subsection{Adding additional MAVs to Collision/Obstacle Avoi- dance Scenario}\label{sec:results2}
The above mission fully demonstrate the essence of the power of the CNMPC. All agents perform individual reference tracking, starting in a tight formation, while avoiding collisions with obstacles and other agents and also keeping the control actions smooth. It is also easy to add additional agents in the obstacle/collision avoidance scenario e.g, the two innermost agents avoid the obstacle and outer agents avoid collisions among each other as the formation is changed due to the obstacle avoidance maneuver. Thus, this configuration will be used as the benchmark test in this article. 

The considered factors are the computation time (mean and maximum) as well as the violation of the constraints for both safety distance and obstacles.
Due to computation times restrictions for additional agents the penalty method iterations are forced to be kept low. In such a way computation time is lowered at a cost of losing the guarantee that no constraints are violated which becomes more apparent as more agents are added. 

\begin{figure}
\centering
  \includegraphics[width=\linewidth]{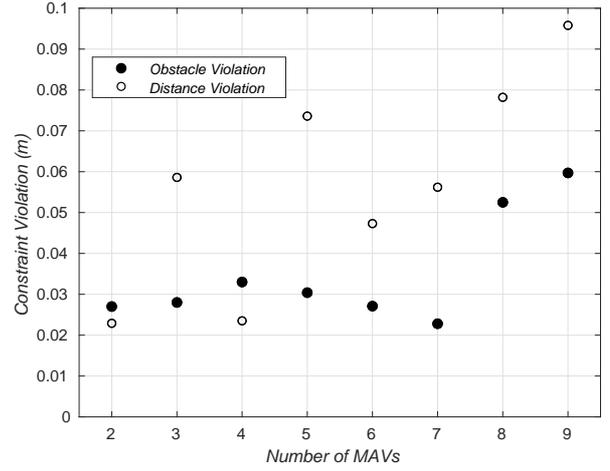}
  \caption{The maximum violation of obstacle and safety distance constraints for the multiple MAV simulations.}
  \label{fig:Constraints}
\end{figure}

Thus, when using OpEn, as a centralized path planner for multiple agents, the behavior shown in Figure \ref{fig:Constraints}, where the maximum constraint violations are shown, is expected. For larger optimization problems, the constraint violations also increase. Table \ref{tab:solvertimes} displays the solver times for the simulation scenarios for additional agents, up to a total of nine, where the computation breaks the threshold of $\unit[50]{ms}$, with seven and eight agents barely above. Moreover, at 8-9 agents Figure \ref{fig:Constraints} shows a fast increasing violation of the constraints. This, in combination with the increase in solver time, shows the limit of the range of OpEn for this specific configuration of constraints, weights and penalty iterations. 

The proposed method show significantly lower computation time compared to centralized and distributed NMPCs solved by Sequential Quadratic Programming (SQP) using \texttt{fmincon} in ~\cite{mansouri2015distributed} for the trajectory planning of multiple MAVs with nonlinear models and constraints for safety distance between the MAVs. The problem is solved with prediction horizon of three, sampling time of $\unit[0.2]{s}$, and only a one step prediction of \gls{mav} positions are shared between the agents. In this SQP implementation the mean of computation time reaches to at least $\unit[47]{s}$ and $\unit[16]{s}$ with more than two agents for centralized and distributed MPCs respectively. 

\begin{figure}
\centering
  \includegraphics[width=\linewidth]{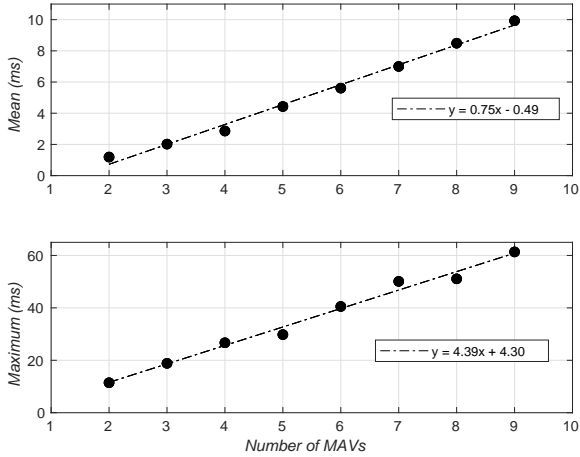}
  \caption{Minimum solver time (top), Average solver time (middle) and Maximum solver time (bottom) for Obstacle/Collision avoidance simulations.}
  \label{fig:SolverTimes}
\end{figure}

The comparison with this article is not completely fair, but it does provide a very interesting result that is demonstrated in Figure \ref{fig:SolverTimes} where additional agents increase both considered solver times linearly, which is the result of a distributed network in \cite{mansouri2015distributed}. In the same article, a centralized scheme results in an exponential increase in solver time for additional agents, again using \texttt{fmincon}. Demonstrated here OpEn shows a different behavior as more decision variables and constraints are added in the form of additional agents.

The aforementioned \citep{kamel2017nonlinear} and \citep{erunsal2019nonlinear} achieve real-time applicable solver times using decentralized and centralized NMPC, while considering two and three agents respectively, but include no analysis for an increasing number of agents and are also very different in their collision avoidance approach. 
{\renewcommand{\arraystretch}{1.3}
\begin{table}
\centering
  \caption{Mean and maximum solver time for simulations of 2--9 agents.}
  \label{tab:solvertimes}
  \begin{tabular}{ c||c|c }
    \hline
   \multirow{2}{*}{\# of MAVs}  & \multicolumn{2}{|c}{Solver Time (ms)} \\
   \cline{2-3} 
      & Mean & Maximum\\
    \hline
    2       & 1.19 &  11.4\\
    3  & 2.02 &  18.8 \\
    4    & 2.86&  26.7\\
    5 & 4.43 &29.8\\
    6  & 5.61   & 40.5\\
    7  & 7.08 &50.0\\
    8  & 8.49 & 51.1\\
    9 & 9.93 & 61.3\\ 
    \hline
  \end{tabular}
\end{table}
}

\subsection{Demanding Collision Avoidance with four MAVs}\label{sec:results3}
\begin{figure}
\centering
  \includegraphics[width=\linewidth]{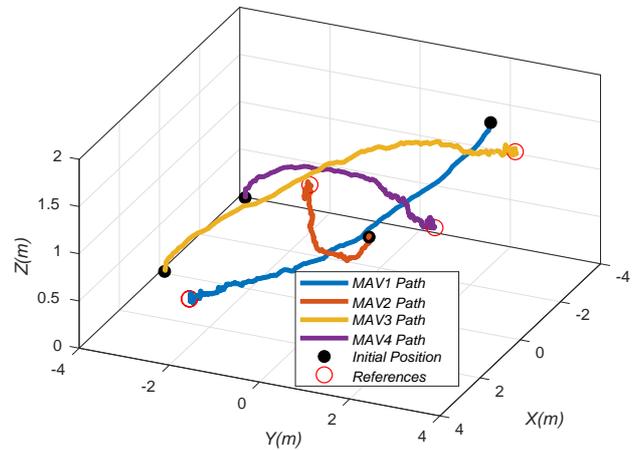}
  \caption{Paths of 4 MAVs maneuvering in 3D while keeping a distance of 0.4m.}
  \label{fig:4AvoidancePath}
\end{figure}

The previous simulations have all considered a kind of formation flight, where all agents are moving in the same direction and the formation is broken by the obstacle avoidance maneuver. In Figure \ref{fig:4AvoidancePath} a scenario where four agents set on a direct collision course with each other is demonstrated. Due to the demanding scenario the penalty method iterations were increased from four to five in this simulation. 

\begin{figure}
\centering
  \includegraphics[width=\linewidth]{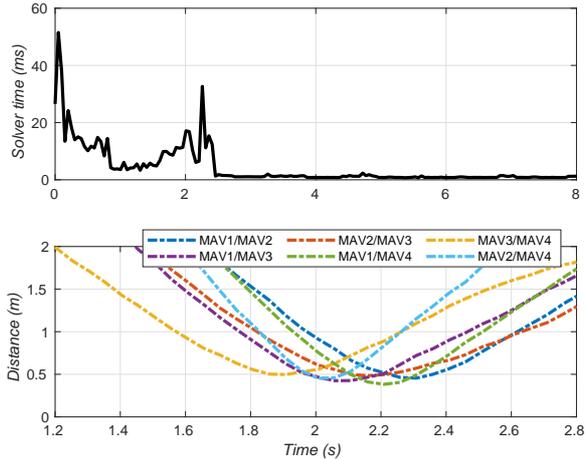}
  \caption{Solver time (top) and distance between agents (bottom) from collision avoidance simulation with 4 MAVs}
  \label{fig:4AvoidanceData}
\end{figure}

In this case, all agents avoid each other, while moving from an initial position to a reference position. Figure \ref{fig:4AvoidanceData} shows the distance between agents, as well as the solver time which peaks at the threshold of 50ms for the second computed time step. The fact that all agents almost precisely reach the allowed distance before separating again, while moving to the reference in such a demanding scenario, is a great result for the collision avoidance. This can be attributed to the nature of the centralized scheme, where the predicted states of all agents are solved for/iterated simultaneously, as to orchestrate trajectories that precisely satisfy the constraints. 


\subsection{Obstacle Course with six MAVs}\label{sec:results4}
\begin{figure}
\centering
  \includegraphics[width=\linewidth]{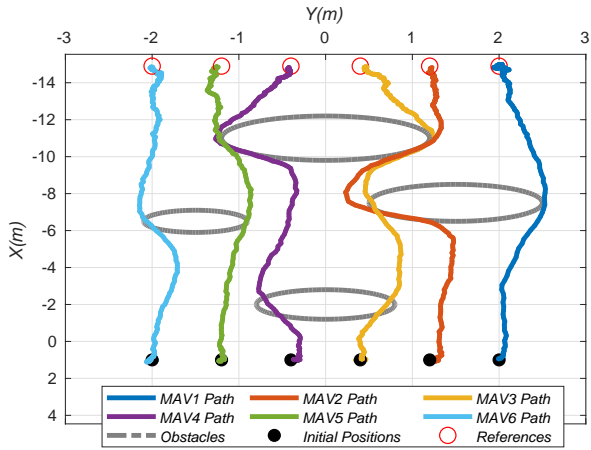}
  \caption{Paths of six MAVs avoiding a series of cylinder obstacles with varying radii while keeping a distance of 0.4m.}
  \label{fig:ObstacleCoursePath}
\end{figure}

Considering the results in Figure \ref{fig:Constraints}, performance starts to deteriorate at around seven or more agents. Thus, to test the method within a reasonable limit, Figure \ref{fig:ObstacleCoursePath} displays a multiple-obstacle scenario using six agents. Another difference compared to simulations in Section \ref{sec:results2} being the longer simulated path. Figure \ref{fig:ObstacleCoursePath} shows a 2D plot for the paths of the agents, since for this number a 3D representation becomes quite chaotic. Six agents move through a constrained environment avoiding multiple obstacles (all cylindrical obstacles with different radii) while keeping a safety distance of minimum $\unit[0.4]{m}$ among agents. 

\begin{figure}
\centering
  \includegraphics[width=0.95\linewidth]{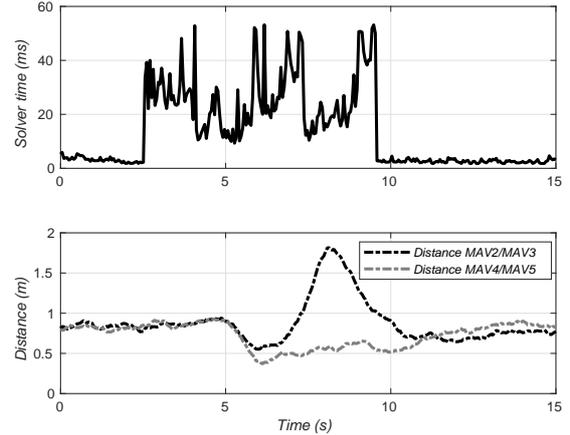}
  \caption{Solver time (top) and Distance between relevant agents (bottom) from six-agent obstacle course simulation.}
  \label{fig:ObstacleCourseData}
\end{figure}

This simulation shows the use of centralized NMPC as a local path planner for a distance much longer than the specified control horizon, smoothly moving past multiple obstacles. Figure \ref{fig:ObstacleCourseData} shows the solver time and distance between relevant agents for the simulation. Again the solver time spikes up to around $\unit[50]{ms}$, slightly violating the threshold at two instants, at demanding points during the simulation. Considering the paths of MAV3 and MAV4 in Figure \ref{fig:ObstacleCoursePath} there is a constraint violation when passing the first cylinder. The violation is approximately $\unit[0.06]{m}$, which is slightly above the six-agent simulation in Figure \ref{fig:Constraints}. Due to the nature of the initial conditions and obstacle avoidance maneuvers, only the distance of MAV2 and MAV3 touches the distance constraints as demonstrated in Figure \ref{fig:ObstacleCourseData}, which shows the two relevant distances between agents. 

\section{Conclusions} \label{sec:conclusion}

This article demonstrated a novel centralized NMPC (CNMPC) using OpEn through multiple simulations. The proposed CNMPC architecture accounted for obstacles and collision avoidance, while keeping the solver time below the considered 50ms threshold for up to seven agents. An analysis of the solver times and the constraint violations for up to nine agents performing simultaneous collision/obstacle avoidance maneuvers is included. The CNMPC successfully completed the missions of set point tracking for multiple agents in demanding environments, while keeping the control signals smooth and without spikes. For the CNMPC framework based on OpEn, the execution time scales linearly for an increasing number of agents, or decision variables, when it comes to solver time. The weakness of the approach of using a penalty method, with a low number of penalty method iterations to keep the solver times low is considered, as the constraint violations also increase with an increasing number of agents, and should be taken into account when applying this approach. Future works include applying this method to laboratory experiments to demonstrate the efficacy of an CNMPC framework for centralized control in a real-world real-time application. 

\bibliography{mybib}
\end{document}